\documentclass[10pt,twocolumn,letterpaper]{article}

\usepackage[pagenumbers]{cvpr} 

\usepackage{pifont}
\usepackage{makecell}

\newcommand{\modelname}{CMD-SE }

\usepackage[dvipsnames]{xcolor}

\definecolor{cvprblue}{rgb}{0.21,0.49,0.74}
\usepackage[pagebackref,breaklinks,colorlinks,citecolor=cvprblue]{hyperref}
\usepackage[accsupp]{axessibility}
\def\paperID{1593} 
\def\confName{CVPR}
\def\confYear{2024}

\title{Exploring the Potential of Large Foundation Models for Open-Vocabulary HOI Detection}

\author{Ting Lei\quad Shaofeng Yin\quad Yang Liu\thanks{Corresponding author} \\
Wangxuan Institute of Computer Technology, Peking University \\
{\tt\small \{ting\_lei, yangliu\}@pku.edu.cn} \quad 
{\tt\small yin\_shaofeng@stu.pku.edu.cn} \\
}

\begin{document}
\maketitle
\begin{abstract}

Open-vocabulary human-object interaction (HOI) detection, which is concerned with the problem of detecting novel HOIs guided by natural language, is crucial for understanding human-centric scenes. 
However, prior zero-shot HOI detectors often employ the same levels of feature maps to model HOIs with varying distances, leading to suboptimal performance in scenes containing human-object pairs with a wide range of distances.
In addition, these detectors primarily rely on category names and overlook the rich contextual information that language can provide, which is essential for capturing open vocabulary concepts that are typically rare and not well-represented by category names alone.
In this paper, we introduce a novel end-to-end open vocabulary HOI detection framework with conditional multi-level decoding and fine-grained semantic enhancement (CMD-SE), harnessing the potential of Visual-Language Models (VLMs). Specifically, we propose to model human-object pairs with different distances with different levels of feature maps by incorporating a soft constraint during the bipartite matching process. 
Furthermore, by leveraging large language models (LLMs) such as GPT models, we exploit their extensive world knowledge to generate descriptions of human body part states for various interactions. Then we integrate the generalizable and fine-grained semantics of human body parts to improve interaction recognition.
Experimental results on two datasets, SWIG-HOI and HICO-DET, demonstrate that our proposed method achieves state-of-the-art results in open vocabulary HOI detection. 
The code and models are available at \url{https://github.com/ltttpku/CMD-SE-release}.

\end{abstract}    
\section{Introduction}
\label{sec:intro}

\begin{figure}[t]
  \centering
   \begin{subfigure}{0.95\linewidth}
        \includegraphics[width=0.95\linewidth]{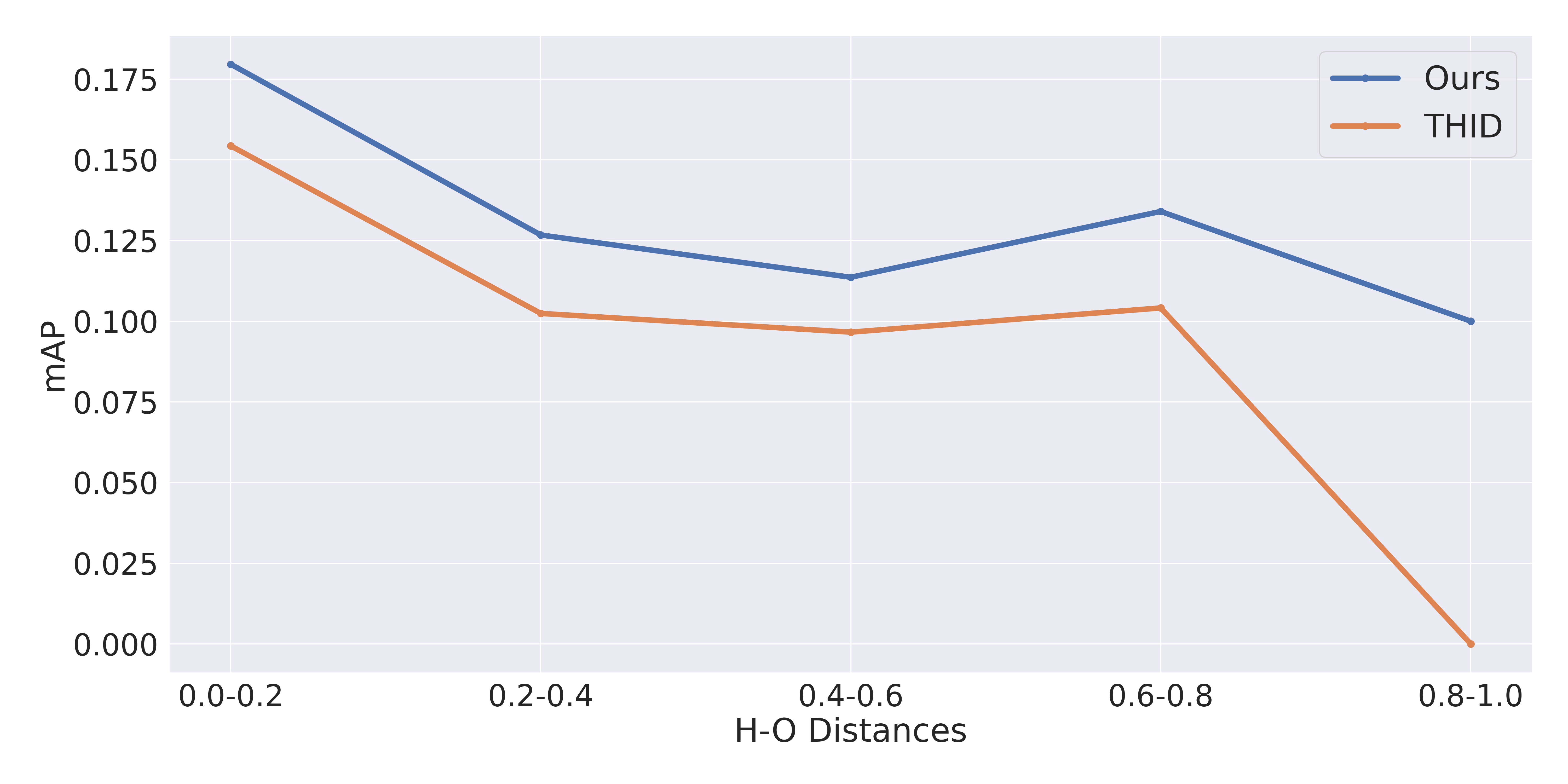}
        \caption{Performance comparison between our method and THID on HOIs with different distances on the open-vocabulary SWIG-HOI dataset.}
        \label{fig:HO-distance}
  \end{subfigure}
  \hfill
  \begin{subfigure}{0.95\linewidth}
        \includegraphics[width=0.95\linewidth]{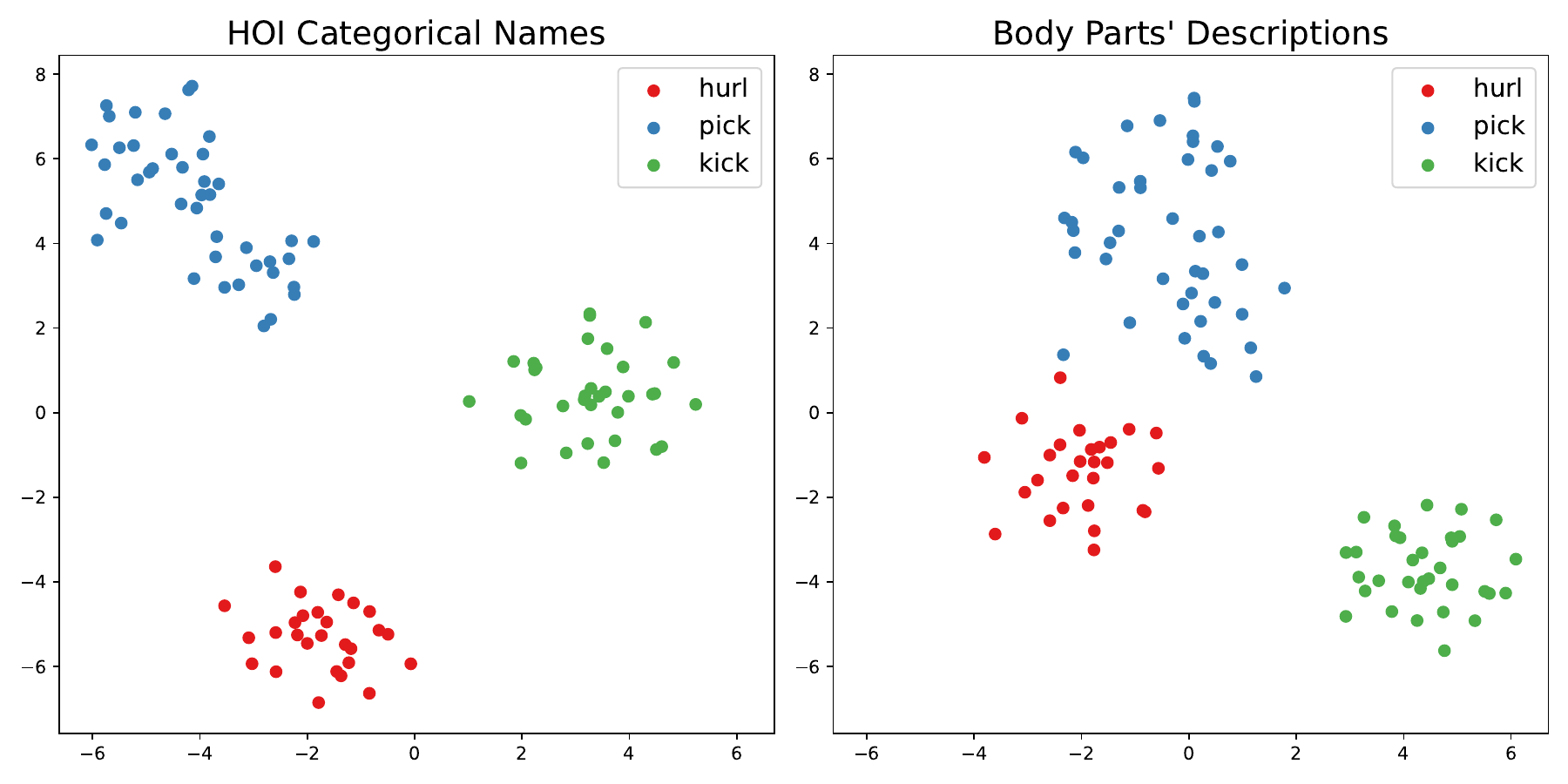}
        \caption{t-SNE visualization of HOI concepts in semantic space. }
        \label{fig:embedtext}
  \end{subfigure}
  
   \caption{(a) Previous method~(THID) suffers from severe performance drop on HOIs with larger distances in the open-vocabulary setting. (b) Compared with HOI categorical names, body parts' descriptions could better recognize the correlation of human postures between different actions. For instance, the action of hurling and picking typically involves extended arms, whereas kicking is characterized by extended legs.}
   
  \label{fig:teasor}
\end{figure}


Human-object interaction (HOI) detection aims to localize the interacting human and object pairs and then recognize their interactions, \textit{i.e.} a set of $<$human, object, action$>$ triplets. 
Most existing HOI detection algorithms~\cite{kim2021hotr,li2022improving,dong2022category,liu2022interactiveness_field,Zhang_2022_STIP,zhang2021CDN,zhang2022UPT,zhang2023pvic,cao2023RmLR,Xie_2023_CVPR} are unscalable in terms of vocabulary size, i.e., they treat interactions as discrete labels and train a classifier based on a predefined category space.
However, due to the combinatorial nature of interactions, it is impractical to create a data collection to include all possible HOIs, especially when the action and object category space becomes large. This motivates us to study a transferable HOI detector that can detect any HOI given its class name.

Recent works~\cite{kato2018compositional,bansal2020func, hou2020VCL,hou2021FCL,hou2021ATL} propose to utilize compositional learning to enhance the generalization ability of HOI detectors, particularly for unseen interactions. Their core idea is to decompose the interaction into action and object, followed by data augmentation to create novel combinations. However, lacking the help of semantics, the above methods can only detect a small set of HOIs predefined in the dataset. Besides, the aforementioned methods may struggle to detect rare concepts, which are more likely to represent new Open Vocabulary (OV) concepts not included in the annotated data during training.
Other works~\cite{liao2022gen,wu2022EoID,ning2023hoiclip,wang2022_THID} propose to incorporate language priors in zero-shot HOI detection, transforming one-hot HOI labels into natural language supervision by CLIP~\cite{CLIP}.

Despite the progress, previous methods still face two major challenges in solving the open-vocabulary HOI detection task.
First, the distances between interactive human-object pairs often exhibit diverse spatial distances. As a result, modeling HOIs using uniform levels of feature maps may lead to suboptimal performance~\cite{kim2022mstr}. As shown in Figure~\ref{fig:HO-distance}, the existing state-of-the-art approach THID suffers from a severe performance drop on HOIs with large distances.
Second, previous detectors rely on category names and overlook the rich contextual information that language can provide. This oversight is particularly pertinent in the context of open vocabulary, where rare concepts are prevalent. In such cases, text embeddings for rare concepts may not be reliable and may fail to reveal the visual appearance similarities between them. Consequently, these detectors may face challenges in capturing the intrinsic connections among different HOI concepts and in generalizing this knowledge from common categories to rare or unseen categories.
For example, considering three actions: hurling, picking, and kicking, we can visually discern that hurling and picking involve extended arms and open hands, while kicking is characterized by extended legs and feet. However, in the semantic space depicted on the left side of Figure~\ref{fig:embedtext}, the distances between the categorical names of these three actions' HOIs are equivalent, undermining the model's capacity to comprehend interactions in an open-vocabulary scenario.

According to the challenges mentioned above, we build an end-to-end open vocabulary HOI detector with large foundation models, which not only models human-object pairs with different distances by utilizing different levels of feature maps, but also exploits the generalizable and fine-grained semantics of human body parts to distinguish various interaction concepts.
\textit{Firstly,} to deal with the problem of suboptimal performance with single-level feature maps, we propose the utilization of multi-level feature maps tailored to HOIs at different distances. Our strategy involves decoding HOIs in parallel from these multi-level feature maps. Specifically, we guide the low-level feature maps to correspond with human-object pairs that have a smaller H-O distance, and the opposite for higher-level maps. This targeted matching is facilitated by introducing an additional constraint during the bipartite matching process~\cite{kuhn1955hungarian}.
\textit{Secondly,}, we propose to utilize the generalizable and fine-grained semantics of human body parts besides category names for interaction recognition. By focusing on body parts, which are more fundamental and elementary compared to the human body as a whole, we aim to enable a more generalizable visual perception across seen and unseen scenarios.
Inspired by the remarkable world knowledge on a variety of topics of large language models (LLMs) like GPT models~\cite{radford2018GPT1,radford2019GPT2,brown2020GPT3}, we utilize it as an implicit knowledge base to exploit the generalizable and fine-grained semantics of human body parts to enhance the understanding of interaction concepts. Specifically, we query a LLM with natural language to let it describe the states of human body parts given a HOI. Then, we utilize the generated descriptions of human body parts for interaction recognition, guiding the model in a fine-grained semantic space. As illustrated on the right side of Figure~\ref{fig:embedtext}, body parts' descriptions can align the two closely related actions~(hurling and picking), which share common patterns in body parts involvement. Simultaneously, this enhancement effectively distinguishes them from kicking, thereby amplifying the model's ability to comprehend a broader range of interactions.

Our contributions are summarized below. 
(1) We propose to utilize different levels of feature maps of VLM to model HOIs with varying distances in the open-vocabulary scenario through a conditional matching mechanism.
(2) We incorporate the generalizable and fine-grained semantics of human body parts obtained by querying LLM to enhance the understanding of a large vocabulary of interactions.
(3) Experiments on two datasets, SWIG-HOI and HICO-DET validate that the proposed method can achieve state-of-the-art results on Open-Vocabulary HOI detection.

\section{Related Work}
\label{sec:related_work}

\subsection{Generic HOI Detection}
According to the network architecture design, previous HOI detection methods can be categorized into two-stage~\cite{cao2023RmLR,zhang2023pvic,zhang2022UPT,gao2020drg,li2019transferable,gao2018ican,Park_2023_CVPR,ting2023hoi} and one-stage~\cite{zhong2021glance,liao2020ppdm,chen2021reformulating,kim2020uniondet,gkioxari2018detecting,kim2022mstr,zhang2021CDN,Tu_2023_ICCV,Kim_2023_CVPR} paradigms. 
Two-stage methods usually apply an object detector first to detect humans and objects, followed by specifically designed modules for human-object association and interaction recognition. They typically use multi-steam~\cite{gupta2019no,li2020pastanet,liu2022interactiveness_field,hou2021FCL} or graph-based~\cite{ulutan2020vsgnet,yang2020graph} methods to support interaction understanding. 
In contrast, one-stage methods typically employ multitask learning to perform instance detection and interactive recognition jointly~\cite{liao2020ppdm,zhang2021CDN,tamura2021qpic,kim2021hotr}.
Despite the progress, standard HOI detection treats interactions as discrete labels and learns a classifier based on a predefined category space, lacking the ability to detect numerous potential unseen interactions.

\subsection{Vision-Language Modeling in HOI Detection}
Although previous HOI detectors have achieved moderate success, they often treat interactions as discrete labels and ignore the richer semantic text information in triplet labels.
More recently, some researchers~\cite{zhong2020polysemy,iftekhar2022SSRT,wang2022_THID,liao2022gen,ning2023hoiclip,wu2022EoID,zhou2019RLIP,kim2022mstr} have investigated the generalizable HOI detector with Vision-Language Modeling.
Among them, PD-Net~\cite{zhong2020polysemy} and SSRT~\cite{iftekhar2022SSRT} propose to aggregate language prior features into the interaction recognition.
RLIP~\cite{zhou2019RLIP} proposes a pre-training strategy for HOI detection based on image captions.
GEN-VLKT~\cite{liao2022gen} and HOICLIP~\cite{ning2023hoiclip} employ the CLIP visual encoder to guide the learning of interaction representation and utilize CLIP text embeddings of prompted HOI labels to initialize the classifier.
THID~\cite{wang2022_THID} proposes a HOI sequence parser to detect multiple interactions and first gets promising results on the recent open-vocabulary HOI detection dataset~\cite{wang2021SWIG-HOI}.
OpenCat~\cite{Zheng_2023_CVPR} leverages massive amounts of weakly supervised data and proposes several proxy tasks for HOI pre-training based on CLIP.
However, previous methods suffer from the limitation of utilizing the same levels of feature maps to model HOIs with varying distances, resulting in suboptimal performance. Additionally, these methods overlook the valuable and generalizable semantics of human body parts, which can enhance the understanding of interactions at a fine-grained level.

\subsection{Leverage LLM for Text Classifier}

Distinct from traditional methods like manually crafting descriptions~\cite{he2017fine,reed2016learning} or utilizing external databases such as Wikipedia~\cite{elhoseiny2017link,naeem2022i2dformer} or the WordNet hierarchy~\cite{fellbaum1998wordnet,roth2022integrating,shen2022k}, recent studies have demonstrated the effectiveness of Large Language Models (LLMs) in generating descriptive prompts for the classification and detection tasks. Some works~\cite{pratt2023does,naeem2023i2mvformer,menon2022visual} utilize GPT-3~\cite{brown2020GPT3} to create detailed sentences that capture visual concepts for recognizing specific categories. Others~\cite{novack2023chils,roth2023waffling} propose using LLMs to generate semantic hierarchies or high-level concepts to improve semantic understanding for zero-shot class prediction tasks.
Recently, many works~\cite{unal2023weaklysupervised,li2023zeroshot, kaul2023multimodal, cao2023detecting, jin2024llms,zang2023contextual} also employ LLM to generate fine-grained descriptions for detection tasks. \cite{zang2023contextual} proposes a novel framework that utilizes contextual LLM tokens as conditional object queries to enhance the visual decoder. \cite{unal2023weaklysupervised} leverages LLM's semantic prior as a filter to filter out HOIs that are unlikely to interact.
Given the value of external knowledge provided by LLMs, it is crucial to investigate how to efficiently integrate this knowledge for a better understanding of HOI concepts. However, investigating how to integrate this knowledge is \textit{underexplored in the HOI} community.
In our work, we explore the LLM-generated descriptions tailored for human-centric relationship understanding. Instead of using the general description of the object or action simply, we propose to utilize the body parts descriptions generated by querying LLM for fine-grained semantic enhancement during interaction recognition. The incorporation of body part descriptions facilitates a more nuanced understanding of interactions, enabling greater generalization across diverse human-centric scenarios.

\section{Method}

\begin{figure*}[t]
  \centering
   \includegraphics[width=0.95\linewidth]{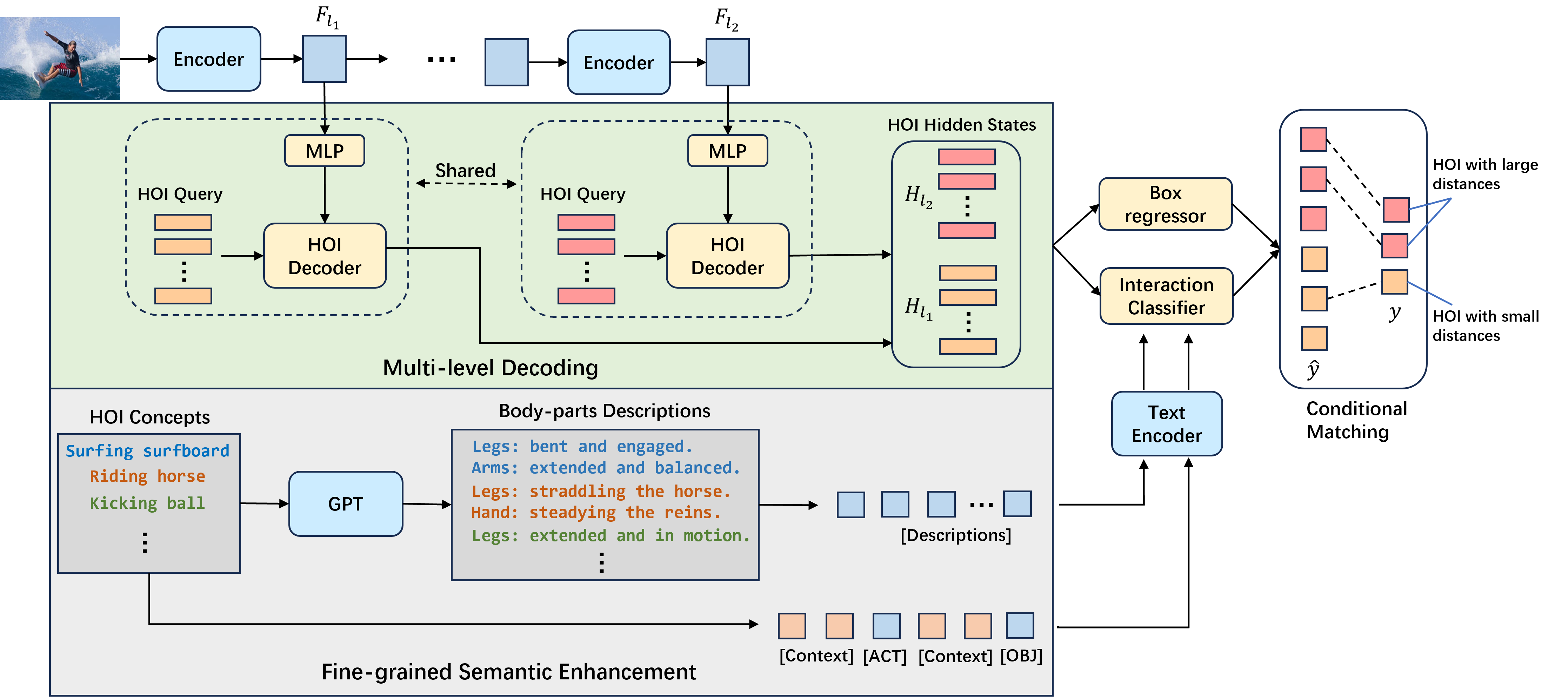}
   \caption{The framework of our CMD-SE. Given an image, the visual encoder is first applied to extract the multi-level visual features. Then we decode the HOIs from multi-level feature maps through a shared HOI decoder parallelly and encourage the HOIs decoded from low-level feature maps to model HOIs with small distances and vice versa via conditional matching. Additionally, we query GPT to describe the states of human body parts for each HOI and utilize the generalizable and fine-grained descriptions as additional prompts to improve interaction recognition.}
   \label{fig:pipeline}
\end{figure*}

\label{sec:method}
In this section, we aim to design an open-vocabulary HOI detector that can detect HOIs described by arbitrary text inputs leveraging large foundation models. 
An overview of our Open-Vocabulary HOI detector is shown in Figure~\ref{fig:pipeline}.
In the following, we start with the formal problem formulation and introduce a basic pipeline with an encoder-decoder architecture in Section~\ref{subsec:preliminary}, harnessing the potential of CLIP~\cite{CLIP}. 
We then introduce the process of multi-level decoding with the conditional matching of our \modelname in Section~\ref{subsec:multi-level}. This design motivates the model to detect HOIs with different distances with different feature maps, alleviating the potential suboptimal issue.
In Section~\ref{subsec:body-parts}, we describe the generation and incorporation of fine-grained semantics pertaining to human body parts behind the interaction labels. This approach enhances the model's ability to discern correlations in human postures across different actions, guiding the model within a more structured semantic space.
At last, we present loss functions used to train our \modelname in Section~\ref{subsec:training}.

\subsection{Preliminary}
\label{subsec:preliminary}

\noindent \textbf{Problem Formulation.}
We define an interaction as a tuple $(b_{o}, b_{h}, o, a)$, where $b_{o}, b_{h} \in \mathbb{R}^{4}$ denote the detected bounding box of a human and object instance, respectively. $o \in \mathbb{O}$ and $a \in \mathbb{A}$ denote the object and action category, where $\mathbb{A} = \{1, 2, ..., A\}$ and $\mathbb{O} = \{1, 2, ..., O\}$ denote the human action and object set, respectively. The objective of open-vocabulary HOI detection is to accurately detect a wide range of interactions. This entails the capability to recognize interactions that have not been encountered during the training phase, encompassing unseen objects, actions, and their various combinations.

\noindent \textbf{A Basic Pipeline.}
We first build an end-to-end open-vocabulary HOI detector with the help of the generalization capability on classification tasks of CLIP~\cite{CLIP} and cast HOI detection as an end-to-end set matching problem similar to DETR~\cite{carion2020DETR}, eliminating the need for handcrafted components like anchor generation.
Given an image $\mathbf{I}$, the global context representation $F_{img}$ is first extracted by a pre-trained CLIP visual encoder $\mathrm{E_V}$:
\begin{equation}
  F_{img} = \mathrm{E_V}(\mathbf{I})
  \label{eq:encodE_Vmage}
\end{equation}
where $F_{img} \in \mathbb{R}^{HW\times C}$ denotes a sequence of feature embeddings of $\mathbf{I}$. We then utilize a transformer-based decoder $D$ to decode the HOIs in $\mathbf{I}$. Specifically, taking the projected context feature and HOI queries $Q = (q_1, q_2, ..., q_M)$ as input, the output from the final layer of $\mathrm{D}$ serves as the representation of interactions:
\begin{equation}
    H = \mathrm{D}(Q, F_{img})
\end{equation}
where $Q$ is treated as query, and the projected context representation is treated as key and value during the cross-attention mechanism of the HOI decoder $\mathrm{D}$, $H = (h_1, h_2, ..., h_M)$, where $M$ denotes the number of HOI queries. Then, we feed them to two different head networks: 1) a bounding box regressor $\mathrm{P_{bbox}}$, which predicts a confidence score $c$ and the bounding box of the interacting human and object $(b_h, b_o)$. 2) a linear projection $\mathrm{P_{cls}}$ which maps the feature to the joint vision-and-text space. Similar to~\cite{wang2022_THID, wu2022EoID, ning2023hoiclip}, we compute its similarity with the features from the text encoder for interaction recognition. Finally we compute the box regression loss $\mathcal{L}_{b}$, the intersection-over-union loss $\mathcal{L}_{iou}$, and the interaction classification loss $\mathcal{L}_{cls}$. Similar to~\cite{liao2022gen}, the matching cost is formulated as:
\begin{equation}
    \mathcal{L}_{cost} = \lambda_b \sum\limits_{i\in\{h,o\}} \mathcal{L}_{b}^{i} \ 
    + \lambda_{iou} \sum\limits_{i\in\{h,o\}} \mathcal{L}_{iou}^{i} \ 
    + \lambda_{cls} \mathcal{L}_{cls}
\label{eq:L_cost}
\end{equation}

\subsection{Conditional Multi-level Decoding}
\label{subsec:multi-level}
Since the variance of the distances between interactive human-object pairs becomes larger as the vocabulary space of interaction grows, we propose using multi-level feature maps to model HOIs with different center distances to avoid suboptimal performance~\cite{kim2022mstr}.
In this subsection, we first formulate the procedure of decoding HOIs from multi-level feature maps. Then we introduce a conditional matching mechanism that encourages different levels of feature maps to model different types of HOIs explicitly, as illustrated in Figure~\ref{fig:pipeline}.

\noindent \textbf{Multi-level Decoding.}
The visual encoder $\mathrm{E_V}$ takes an image $\mathbf{I}$ with a fixed resolution (e.g., 224 × 224), divides it into small patches, and first projects them as a sequence.
Assume $\mathrm{E_V}$ = ($\mathrm{E_V^0}$, $\mathrm{E_V^1}$, ..., $\mathrm{E_V^N}$), where $\mathrm{E_V^i}$ is the $i$-th block of $\mathrm{E_V}$. Denote the input as $X_0$, $\mathrm{E_V}$ encodes the image via the self-attention mechanism~\cite{vaswani2017attention} in each layer:
\begin{equation}
    X_{i+1} = \mathrm{E_V^i}(X_i)
\end{equation}
 where $X_i$ denotes the encoded feature map after the i-th block of $\mathrm{E_V}$. Then we utilize multi-level feature maps to decode HOIs in parallel:
\begin{equation}
    H_{l_i} = \mathrm{D}(Q, X_{l_i}), l_i \in \mathbb{L} 
\end{equation}
where $\mathbb{L}$ denotes the set of levels of interest. Then the decoded hidden states of HOIs ${H_{l_i}}$~$(l_i \in \mathbb{L})$ are fed to the bounding box regressor and the interaction classifier to predict the bounding boxes of human-object pairs and the interaction categories, respectively.

\noindent \textbf{Conditional Matching.}
During the bipartite matching~\cite{carion2020DETR} between the set of $N$ predictions $\hat{y}$ and the set of ground truth HOIs $y=\{y_i^K\}$ per image, we design a unique loss to explicitly encourage $H_i$ decoded from the low-level feature map to model HOIs with relatively small center distances and vice versa to alleviate the performance gap across HOIs with different distances. 
Specifically, besides the typical matching loss $\mathcal{L}_{cost}$ introduced in Equation~\ref{eq:L_cost}, we design a novel constraint loss $\mathcal{L}_{d}$ between the normalized level index of $\hat{y_i}$ and the H-O distance in $y_j$:
\begin{equation}
    \mathcal{L}_{d} = \left\| (Lv(\hat{y_i}), g(y_j)) \right\|_1
\end{equation}
where $Lv(\hat{y_i}) \in (0, 1)$ denotes the normalized level index of $\hat{y_i}$, $g(y_j)$ denotes $L_2$ distance between the center of the annotated human and object in $y_j$. 
This design encourages the model to categorize HOIs with distinct distances into separate groups and address HOIs within each group by utilizing a specific feature map tailored to that group's characteristics.

\subsection{Fine-grained Semantic Enhancement}
\label{subsec:body-parts}
To enhance the differentiation of various HOI concepts that can be challenging to distinguish based solely on their category names, we propose to leverage the underlying states of human body parts associated with the interaction labels. The descriptions of these states serve to provide a more fine-grained and comprehensive understanding of HOI concepts.
In this subsection, we present the process of generating descriptions for the states of human body parts involved in HOIs via prompting with GPT. Additionally, we introduce how the semantics of human body parts are integrated to our model to improve interaction recognition.

\begin{figure}[t]
  \centering
    \includegraphics[width=0.95\linewidth]{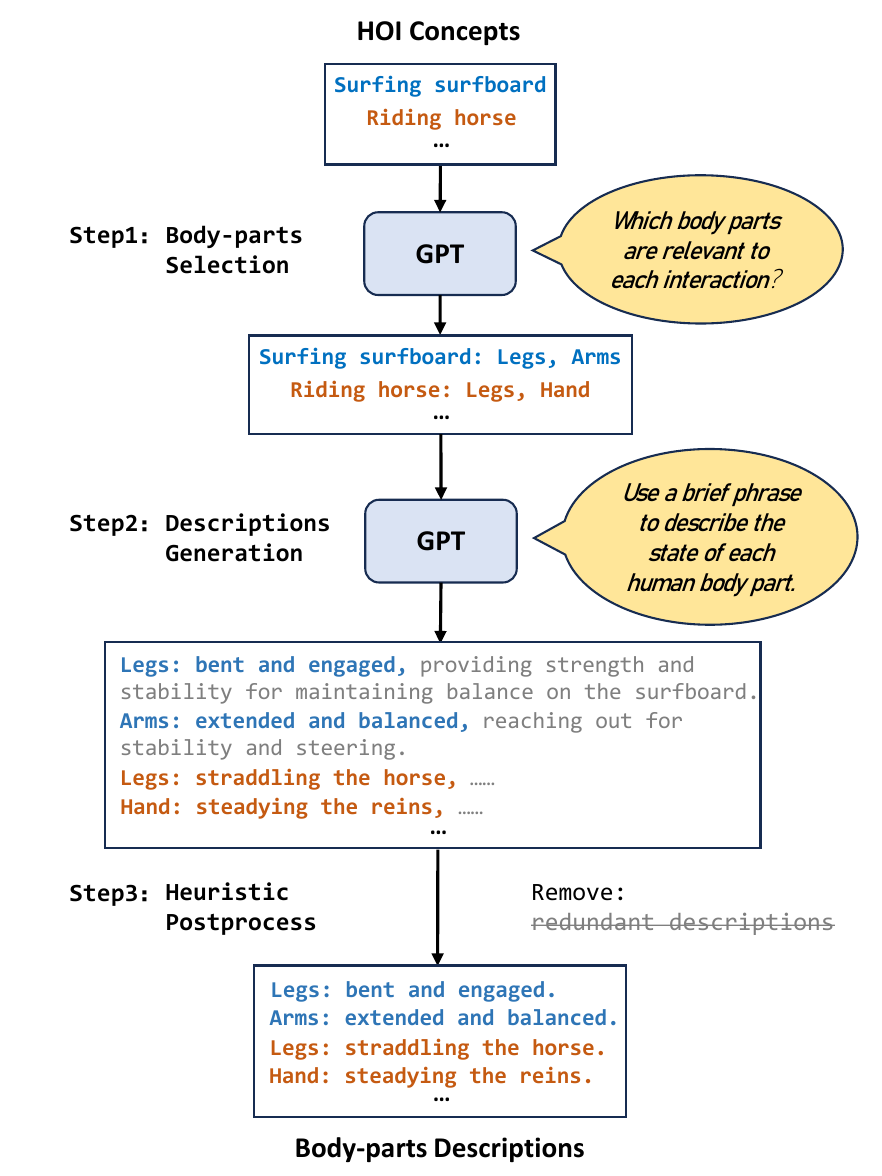}
        \caption{Illustration of generating prompts with GPT. The main purpose of the entire process is to find the most simple and general body parts descriptions for each HOI.}
        \label{fig:gptprompt}
\end{figure}

\noindent \textbf{Prompt with GPT.}
To harness the generalizable and recombinant nature of human body parts for interaction recognition, we employ GPT~\cite{brown2020GPT3} to produce state descriptions of body parts. However, directly querying GPT for body part descriptions may yield verbose descriptions, neglecting the varying relevance of different body parts in interactions. Consequently, we implement a two-step mechanism to query GPT for generating body part descriptions for each HOI.
As illustrated in Figure~\ref{fig:gptprompt}, our approach involves utilizing a standardized pipeline and set of templates as language commands to interact with GPT for each HOI category. Specifically, given an HOI name consisting of an action name $act_i$, and an object name $obj_i$, we initiate the query to GPT by asking: ``Which body parts are involved in the interaction when a person $act_i$ a/an $obj_i$?" This command aids in identifying a subset of body parts $BP_i$ that are pertinent to the current $i$-th interaction:
\begin{equation}
    BP_i = GPT(BP, Query_i)
\end{equation}
where $BP$ represents a predefined list of human body parts, including ``mouth, eye, arm, hand, leg, foot". For example, if the interaction is ``surfing surfboard'', the returned $BP_i$ from GPT would include arm and leg. The main purpose of this step is to limit the description to a few important body parts. If it is generated directly without adding this restriction, the description will not be able to distinguish between key body parts and non-key body parts. This will make it difficult for us to accurately identify the characteristics of the action state and cause confusion in HOI concepts.

Then we further query GPT with the prompt $Query_{i}'$: ``Use a brief phrase to describe the state of each human body part in $BP_i$ when a person $act_i$ a/an $obj_i$.'' to acquire the state descriptions $S_i$ of the corresponding human body parts:
\begin{equation}
\label{eq:body_parts_desctiptions}
    S_i = GPT(BP_i, Query_{i}')
\end{equation}
Finally, we use a heuristic postprocess to remove some redundant descriptions, making the status description more simple and general. 

\noindent \textbf{Enhance Interaction Recognition.}
The primary objective of interaction recognition is to convert the textual description of interactions into a feature space and compare them with the output generated by the HOI decoder in Section~\ref{subsec:multi-level}. In order to address the limited distinctions among HOI names, we also propose leveraging the semantic information provided by descriptions of body parts associated with each HOI. This means that for each HOI, we encode not only its name but also the states of the relevant human body parts.
Following~\cite{zhou2022CoOp,wang2022_THID}, given a HOI category defined as an action-object pair, denoted as $(act_i, obj_i)$, we form a sentence using global learnable context tokens $Ctx$. As illustrated in Figure~\ref{fig:pipeline}, we use a few prefix tokens $Ctx_{pre}$ at the beginning of the sentence and a few learnable conjunct tokens $Ctx_{con}$ to automatically determine how to connect the category names of the action and object. Then, we generate the text embedding for each HOI label through the pre-trained CLIP text-encoder $\mathrm{E_T}$:
\begin{equation}
    T_{hoi} = \mathrm{E_T}(Ctx_{pre}, act_i, Ctx_{con}, obj_i)
\end{equation}
The resulting embeddings of all HOI names are denoted as $T_{hoi} \in \mathbb{R}^{N\times C_t}$, where $N$ denotes the number of HOI triplet categories, and $C_t$ represents the dimension of text embedding from CLIP text encoder.

Furthermore, as the descriptions of body parts are already semantically rich, they do not require learnable vectors as context. To encode the state descriptions $S_i$ of the human body parts for each HOI, as obtained in Equation~\ref{eq:body_parts_desctiptions}, we directly utilize the CLIP text encoder $\mathrm{E_T}$:
\begin{equation}
    T_{b} = \mathrm{E_T}(S_i)
\end{equation}
where $T_{b} \in \mathbb{R}^{N\times C_t}$ represents the embeddings of body parts' descriptions for all HOIs. Note that we only need to forward the body parts descriptions once since there are no learnable parameters in this branch, which significantly enhances its efficiency.
The final interaction prediction is obtained by combining the output logits of $T_{hoi}$ and $T_{b}$:
\begin{equation}
    p_j = \alpha_{hoi} h_j T_{hoi}^T +  \alpha_{b}  h_j T_{b}^T
\end{equation}
where $h_j$ represents the hidden state of a HOI, $p_j$ represents the corresponding interaction prediction, $\alpha_{hoi}$ and $\alpha_{b}$ are learnable scalars used to weigh the contributions of the HOI embeddings~($T_{hoi}$) and body part embeddings~($T_{b}$) respectively.

\subsection{Training and Inference}
\label{subsec:training}
In this subsection, we elaborate the processes of training and inference of our model.

\noindent \textbf{Training.}
During the training stage, we follow the query-based methods~\cite{carion2020DETR,tamura2021qpic,wang2022_THID,liao2022gen} to assign a bipartite matching prediction with each ground-truth using the Hungarian algorithm~\cite{kuhn1955hungarian}.
The matching cost $\mathcal{L}_{cost}$ for the matching process and the targeting cost for the training back-propagation share the same strategy, which is formulated in Section~\ref{subsec:preliminary}. Then considering the additional constraint loss $\mathcal{L}_{d}$ introduced in Section~\ref{subsec:multi-level}, the final matching loss can be defined as:
\begin{equation}
    \mathcal{L} = \mathcal{L}_{cost} + \lambda_{d} \mathcal{L}_{d}
\end{equation}
where $\lambda_{d}$ is the hyper-parameter weight for the additional constraint.

\noindent \textbf{Inference.}
For each HOI prediction, including the bounding-box pair $(\hat{b_{h}^{i}}, \hat{b_{o}^{i}})$, the bounding box score $\hat{c_{i}}$ from the box regressor, and the interaction score $\hat{s_{i}}$ from the interaction classifier, the final score $\hat{s_{i}}'$ is computed as:
\begin{equation}
    \hat{s_{i}}' = \hat{s_{i}} \cdot \hat{c_{i}}^{\gamma}
\end{equation}
where $\gamma$ $\textgreater$ 1 is a constant used during inference to suppress
overconfident objects~\cite{zhang2021scg,zhang2022UPT}.

\section{Experiment}
\label{sec:experiment}

\subsection{Experimental Setting}
\label{subsec:exp_setting}

\noindent \textbf{Datasets.}
Our experiments are mainly conducted on two datasets, SWIG-HOI~\cite{wang2021SWIG-HOI} and HICO-DET~\cite{chao2018HICO-DET}.
The SWIG-HOI dataset encompasses 400 human actions and 1000 object categories.
The test set of SWIG-HOI comprises approximately 14,000 images and 5,500 interactions, with around 1,800 interactions not present in the training set.
The HICO-DET dataset provides 600 combinations involving 117 human actions and 80 objects. We follow~\cite{hou2020VCL,wang2022_THID} to simulate a zero-shot detection setting by holding out 120 rare interactions from all 600 interactions.

\noindent \textbf{Evaluation Metric.}
We follow the settings of previous works~\cite{chao2018HICO-DET,liu2022interactiveness_field,liao2022gen,wang2022_THID} to use the mean Average Precision (mAP) for evaluation. We define an HOI triplet prediction as a true-positive example if the following criteria are met: 1) The IoU of the human bounding box and object bounding box are larger than 0.5 \textit{w.r.t.} the GT bounding boxes; 2) the predicted interaction category is accurate.

\noindent \textbf{Implementation Details.}
Our model is built upon the pretrained CLIP and all its parameters are frozen during training. We employ the ViT-B/16 version as our visual encoder following~\cite{wang2022_THID}. We set the cost weights $\lambda_b$, $\lambda_{iou}$, $\lambda_{cls}$ and $\lambda_{d}$ to 5, 2, 5, and 5 during training and train our model for 80 epochs with a batch size of 128 on 2 A100 GPUs.


\subsection{Comparison with Other Methods}
\label{subsec:compare_sota}
We compare the performance of our model with existing methods on SWIG-HOI and HICO-DET datasets. 
As shown in Table~\ref{tab:swig-hoi}, our model significantly outperforms the previous state-of-the-art on the SWIG-HOI dataset on all splits, achieving a relative gain of 15.08\% on all interactions.
This shows the strong capability of our \modelname to detect and recognize the interactions of human-object pairs in the open-vocabulary scenario.

We also compare our method with state-of-the-art methods on the simulated zero-shot setting of the HICO-DET dataset in Table~\ref{tab:hico-det}. Note that earlier methods~\cite{hou2020VCL,hou2021FCL} lack semantic understanding of interaction categories and predetermine the unseen
interactions during the training stage. Although recent zero-shot methods~\cite{liao2022gen,ning2023hoiclip} employ CLIP text embeddings for interaction classification, they typically rely on a DETR architecture and pretrained DETR weights~(pretrained on object detection datasets with finite categories), which is unscalable in terms of the vocabulary size, thereby limiting their applicability in open-world scenarios. 
Note that it's not a fair comparison between our method and theirs on HICO-DET, as the HICO-DET dataset and the COCO dataset used for DETR pretraining share the same object label space.
In contrast, open-vocabulary methods~\cite{wang2022_THID} break free from this constraint and do not pretrain on any detection datasets. In comparison to the previous open-vocabulary HOI detector, our \modelname achieves new state-of-the-art performance on unseen HOIs.

\begin{table}
  \centering
  \begin{tabular}{@{}lcccc@{}}
    \toprule
    Method & Non-rare & Rare & Unseen & Full \\
    \midrule
    QPIC~\cite{tamura2021qpic} &   16.95   &   10.84   &   6.21    &    11.12   \\
    THID~\cite{wang2022_THID}  &   17.67   &   12.82   &   10.04   &    13.26   \\
    \modelname (Ours)           &  \textbf{21.46} & \textbf{14.64} & \textbf{10.70} & \textbf{15.26} \\
    \bottomrule
  \end{tabular}
  \caption{Comparison of our proposed \modelname with state-of-the-art methods on the SWIG-HOI dataset.}
  \label{tab:swig-hoi}
\end{table}

\begin{table}
  \centering
  \begin{tabular}{@{}lcccc@{}}
    \toprule
    Method & {\makecell[c]{Pretrained \\ Detector}} & Unseen & Seen & Full \\
    \midrule
    \multicolumn{4}{@{}l@{}}{\textit{Zero-shot Methods}} \\
    \hline
    VCL~\cite{hou2020VCL}   & \ding{51}  & 10.06 & 24.28 & 21.43 \\ 
    ATL~\cite{hou2021ATL}   & \ding{51}  & 9.18  & 24.67 & 21.57 \\
    FCL~\cite{hou2021FCL}   & \ding{51}  & 13.16 & 24.23 & 22.01 \\
    GEN-VLKT~\cite{liao2022gen}  &  \ding{51} & 21.36  & 32.91 & 30.56 \\
    HOICLIP~\cite{ning2023hoiclip} & \ding{51} & \textbf{23.48}  & \textbf{34.47} & \textbf{32.26} \\
    \midrule
    \multicolumn{4}{@{}l@{}}{\textit{Open-vocabulary Methods}} \\
    \hline
    THID~\cite{wang2022_THID}& \ding{55}  & 15.53 & \textbf{24.32} & \textbf{22.38} \\
    \modelname (Ours) & \ding{55} & \textbf{16.70} & 23.95 & 22.35 \\
    \bottomrule
  \end{tabular}
  \caption{Comparison of our proposed \modelname with state-of-the-art methods on HICO-DET under the simulated zero-shot setting.}
  \label{tab:hico-det}
\end{table}

\begin{table}
  \centering
  \begin{tabular}{@{}ccccc@{}}
    \toprule
      Method & Non-rare & Rare & Unseen & Full \\
    \midrule
    \textit{Base}  & 15.69 & 11.53 & 7.32 & 11.45 \\
    + \textit{MD}  & 18.66 & 12.93 & 7.29 & 12.87 \\
    + \textit{CM}  & 21.33 & 14.26 & 9.32 & 14.69 \\
    + \textit{SE} & \textbf{21.46} & \textbf{14.64} & \textbf{10.70} & \textbf{15.26} \\
    \bottomrule
  \end{tabular}
  \caption{Ablations of different modules of our \modelname on the SWIG-HOI dataset. \textit{MD}: multi-level decoding. \textit{CM}: conditional matching. \textit{SE}: semantic enhancement.}
  \label{tab:ablation-module}
\end{table}

\subsection{Ablation Study}
\label{subsec:ablation}

\noindent \textbf{Network Architecture.}
In this section, we empirically investigate how the performance of the proposed method is affected by different model settings on the SWIG-HOI dataset.
We denote the basic pipeline model introduced in Section~\ref{subsec:preliminary} as \textit{Base} in Table~\ref{tab:ablation-module}. 
The first modification is applying multi-level features for HOI decoding. We observe an improvement of 1.42\% mAP. Subsequently, we apply the soft constraint to perform conditional matching, denoted as +\textit{CM}, explicitly guiding different levels of feature maps to represent HOIs with varying distances. The results show that +\textit{CM} brings a significant improvement across all splits, contributing to a 2.4\% mAP gain on all categories.
To further enhance the model's semantic understanding of interaction concepts, we incorporate the fine-grained semantics of human body parts, denoted as +\textit{SE}. This modification yields the best performance, with a notable relative gain of 22.8\% in the mAP for unseen classes. This outcome underscores the substantial potential of our framework in enhancing interaction understanding within an open-vocabulary scenario.

\noindent \textbf{Distance Types and Matching Strategies.}
In this section, we explore the impact of different H-O distance types and matching strategies on model performance. Initially, we examine the effectiveness of absolute versus relative distances for matching. While intuition may suggest relative distance would be more effective, as it encapsulates inherent HOI properties, our results, depicted in the first two lines of Table~\ref{tab:ablation-correspondence}, empirically favor absolute distance. We attribute this to its alignment with ViT's attention mechanism, where each patch's attention expands outward by a distance corresponding to the absolute distance in the image. 
As shown in Table~\ref{tab:ablation-correspondence}, matching low-level features with short-distance HOIs achieves superior results compared to matching low-level features for long-distance HOIs, with an improvement of 0.45 and 1.33 mAP on seen and unseen classes, respectively. This finding suggests that low-level features are more effective in capturing interactions that occur in relatively close proximity.


\begin{figure*}
  \centering  
    \begin{subfigure}{0.24\linewidth}
    \includegraphics[width=0.99\linewidth ]{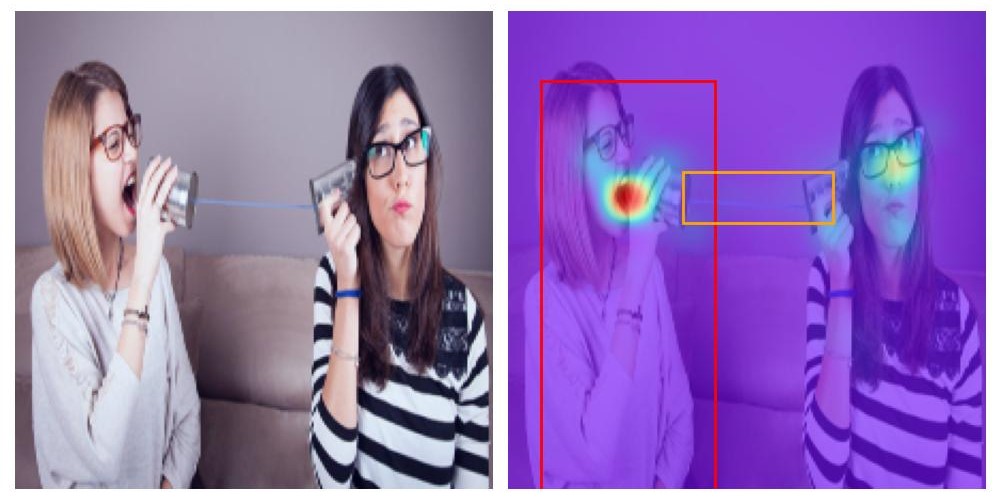}
    \caption{communicating.}
    \label{fig:short-a}
  \end{subfigure}
   \hfill
  \begin{subfigure}{0.24\linewidth}
    \includegraphics[width=0.99\linewidth ]{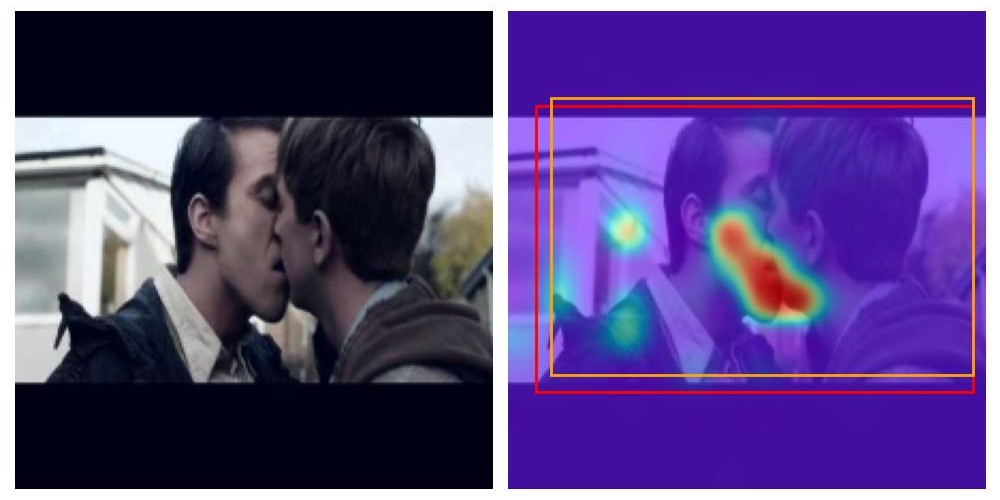}
    \caption{kissing.}
    \label{fig:short-b}
  \end{subfigure}
   \hfill
  \begin{subfigure}{0.24\linewidth}
    \includegraphics[width=0.99\linewidth ]{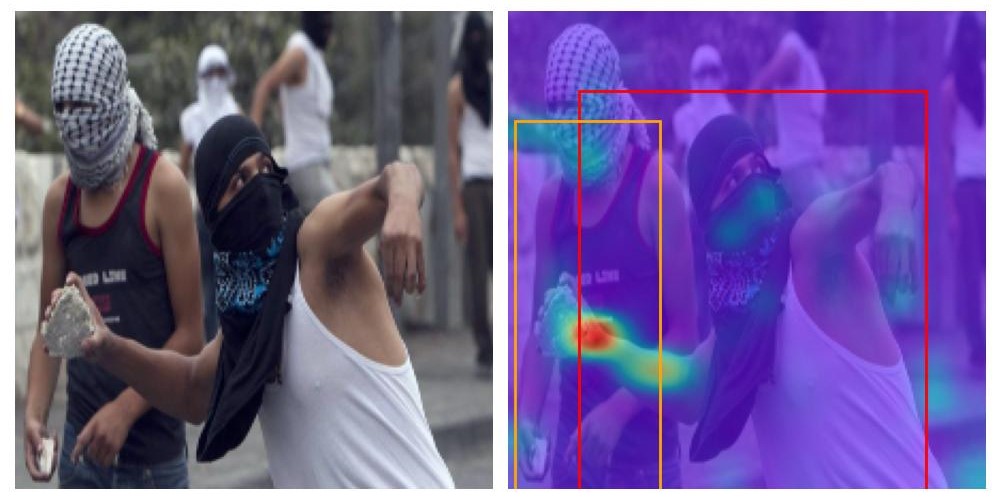}
    \caption{hurling.}
    \label{fig:short-c}
  \end{subfigure}
   \hfill
  \begin{subfigure}{0.24\linewidth}
    \includegraphics[width=0.99\linewidth ]{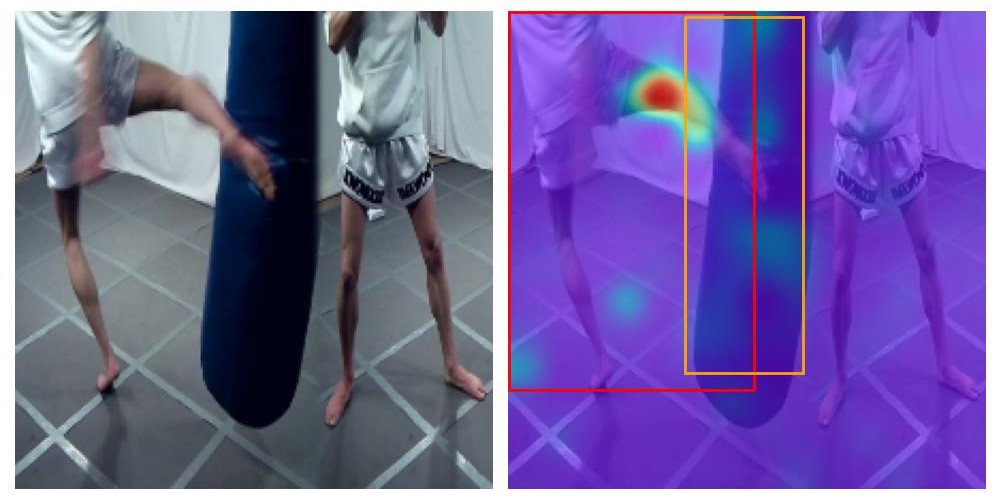}
    \caption{kicking.}
    \label{fig:short-d}
  \end{subfigure}
  
  \caption{Qualitative results of our method on SWIG-HOI test set.}
  \label{fig:short}
\end{figure*}

\noindent \textbf{Additional Constraint for Conditional Matching.}
We discuss the weight of the additional constraint during bipartite matching. As shown in Table~\ref{tab:ablation-L-d}, when setting the weight to 0, the model is optimized to detect HOIs with different distances with the same set of feature maps, leading to suboptimal performance due to the variation in distances between interactive humans and objects.
Setting the weight to 5 encourages our model to differentiate and model HOIs with different distances using distinct feature maps, leading to the best performance and a notable 1.46 mAP gain.
However, increasing the weight further to 10 causes a marginal decline in the model's performance. This can be attributed to the higher weight potentially causing the model to neglect other pertinent factors during bipartite matching.

\noindent \textbf{Prompts for Fine-grained Semantic Enhancement.}
In this part, we explore different types of prompts for enhancing fine-grained semantics. 
A naive and straightforward approach is to describe the action and object associated with each HOI, termed as ``HOI Descriptions''. As shown in the first line of Table~\ref{tab:ablation-prompts}, utilizing HOI descriptions leads to a mAP of 16.92 and 10.52 on seen and unseen classes, respectively. 
Considering the generalizable and recombinant nature of human body parts, we further explore employing body parts for semantic enhancement. Simply utilizing the names of body parts relevant to each HOI, generated by GPT following the ``Body-part Selection'' procedure in Figure~\ref{fig:gptprompt}, results in a marginally inferior performance. This is attributed to the finite set of body part combinations and the presence of the same set of body part names across different HOIs, potentially causing confusion in the model's ability to distinguish them.
By further incorporating descriptions of the states of body parts—characteristics that are not only generalizable and recombinant but also specific to each HOI, our model achieves the best performance. 


\begin{table}
  \centering
  \begin{tabular}{@{}cccc@{}}
    \toprule
    Distance Types & Matching Strategies & Seen & Unseen \\
    \midrule
    Relative &  Low-Small   & 16.57 &  10.30 \\
    Absolute &  Low-Small   & \textbf{16.79} & \textbf{10.70} \\
    Absolute &  Low-Large   & 16.61 & 10.47  \\
    
    \bottomrule
  \end{tabular}
  \caption{Ablations on the H-O distance types and matching strategies between the levels of feature maps and the H-O distances. \textit{Relative}: using relative distance as H-O distance. \textit{Absolute}: using absolute distance as H-O distance. \textit{Low-Small}: matching low layer with small distance HOI.  \textit{Low-Large}: matching low layer with large distance HOI. }
  \label{tab:ablation-correspondence}
  \vspace{-0.5em}
\end{table}


\begin{table}
  \centering
  \begin{tabular}{@{}ccccc@{}}
    \toprule
    $\lambda_{d}$ & Non-rare & Rare & Unseen & Full \\
    \midrule
    0  &  19.80 & 13.94  & 7.89  & 13.80 \\
    5  &  \textbf{21.46} & \textbf{14.64} & \textbf{10.70} & \textbf{15.26} \\
    10 &  20.18 & 14.06  & 9.64  & 14.40 \\
    \bottomrule
  \end{tabular}
  \caption{Ablations of the weight for the additional soft constraint during conditional matching.}
  \label{tab:ablation-L-d}
  \vspace{-1.1em}
\end{table}

\begin{table}
  \centering
  \begin{tabular}{@{}lcc@{}}
    \toprule
    Prompts & Seen & Unseen \\
    \midrule
    HOI Descriptions        & 15.73 & 10.10 \\
    Body Parts Names        & 16.08 & 9.81  \\
    Body Parts Descriptions & \textbf{16.79} & \textbf{10.70} \\ 
    \bottomrule
  \end{tabular}
  \caption{Ablations on different types of prompts for fine-grained semantic enhancement.}
  \label{tab:ablation-prompts}
  \vspace{-1em}
\end{table}

\subsection{Qualitative Results}
\label{subsec:qualitative}
We visualize the prediction result and attention maps to illustrate the characteristics of our method. As shown in Figure~\ref{fig:short-a} and \ref{fig:short-b}, we find that our model can adaptively use the output of different layers to predict HOIs with different human-object pairs distances. For a farther interaction of communicating person and a closer interaction of kissing person, their optimal predictions are encoded by the output of the 8th layer and the 5th layer of the encoder respectively. This shows that our model has generalization ability in different human-object pairs distances. Additionally, as shown in Figure~\ref{fig:short-c} and \ref{fig:short-d}, we find that our model exhibits some sensitivity to human body parts. For these two examples, actions of hurling and kicking, our \modelname focuses on the hands and arms for hurling, and the legs for kicking. This shows that our model can use the characteristics of human body parts to infer the HOI interaction through our design.

\section{Conclusion}
\label{sec:conclusion}

We utilize large foundation models to build an end-to-end open vocabulary HOI detector that addresses two major challenges in the field. First, we tackle the issue of diverse spatial distances between interactive human-object pairs by utilizing different levels of feature maps tailored to HOIs at different distances. 
Second, we leverage fine-grained semantics of human body parts, in addition to category names, to enhance interaction recognition. By querying a LLM for descriptions of human body parts given a HOI, the detector gains a deeper understanding of the correlation of human postures between different actions. Experimental results demonstrate the superior performance of our model in open-vocabulary HOI detection.

\noindent \textbf{Acknowledgements.}
This work was supported by the grants from the National Natural Science Foundation of China 62372014.

{
    \small
    \bibliographystyle{ieeenat_fullname}
    \bibliography{main}
}


\end{document}